# Optical machine learning with incoherent light and a single-pixel detector

SHUMING JIAO, JUN FENG, YANG GAO, TING LEI*, ZHENWEI XIE, XIAOCONG YUAN*

*Nanophotonics Research Center, Shenzhen University, Shenzhen, Guangdong, 518060, China*
*Corresponding author: leiting@szu.edu.cn , xcyuan@szu.edu.cn

**An optical diffractive neural network (DNN) can be implemented with a cascaded phase mask architecture. Like an optical computer, the system can perform machine learning tasks such as number digit recognition in an all-optical manner. However, the system can only work under coherent light illumination and the precision requirement in practical experiments is quite high. This paper proposes an optical machine learning framework based on single-pixel imaging (MLSPI). The MLSPI system can perform the same linear pattern recognition task as DNN. Furthermore, it can work under incoherent lighting conditions, has lower experimental complexity and can be easily programmable. © 2019 Optical Society of America**

http://dx.doi.org/10.1364/OL.99.099999

The research of machine learning and artificial intelligence has received much attention in recent years. A machine learning system is usually realized by digital algorithms on a computer. However, recent works [1-3] demonstrate that machine learning can be implemented physically in all-optical manner with a diffractive neural network (DNN) system. All-optical machine learning systems have some potential advantages over digital ones, such as light-speed computing, parallel processing, and low power consumption [1]. The DNN system [1] consists of several layers of cascaded phase-only masks placed in parallel, which are perpendicular to the light propagation direction, shown in Fig. 1(a). The input light field is sequentially modulated by each different phase mask in the forward free-space propagation. Each layer of phase mask in the system can be analogized to one layer of neurons in a deep neural network system. Consequently, the system exhibits some machine learning capabilities. Similar to the error back-propagation mechanism in digital deep learning, the phase masks can be optimized based on training samples. After training, the designed system can perform certain artificial intelligent tasks such as number digit recognition, like an optical computer. The current configuration in the work [1] is basically still a linear system even though non-linear operations can be possibly implemented in future works [1,3]. Mathematically, all the phase masks jointly perform a certain complex-amplitude linear transform from the input light field to the output target light field.

Despite the success, the proposed DNN system [1] has some drawbacks and limitations. One major limitation is that the system is based on optical diffraction and can only work under coherent light illumination (e.g. laser illumination). In addition, the multiple phase masks need to be aligned very precisely in the experimental setup. Consequently, such a system has a relative high complexity in the practical implementation and optical experiments.

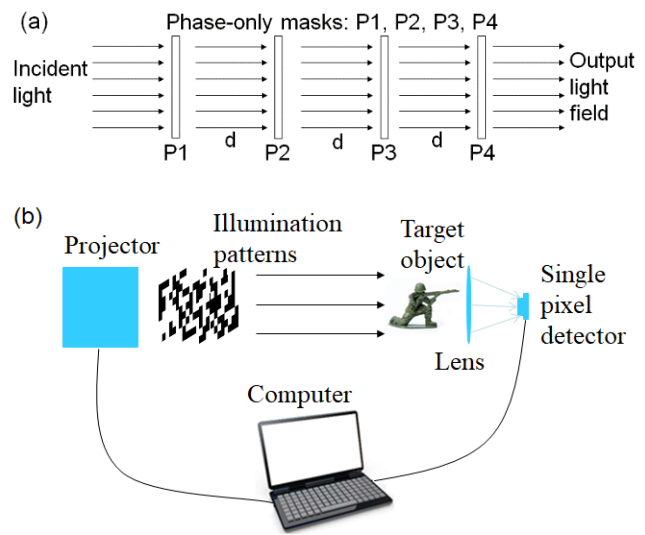

**Fig. 1.** (a)Optical setup for a diffractive neural network (DNN) or a cascaded phase mask system (d is the separation distance between neighboring masks); (b)Optical setup of a signal-pixel imaging system.

In this paper, a single-pixel imaging (SPI )system is proposed to perform the same linear machine learning task as the DNN system (referred to as MLSPI), with advantages of incoherent light illumination and easier experimental realization. In a conventional optical imaging system, the light field is captured by a pixelated sensor array. However, in SPI, the sensor only contains one single pixel. The object image is sequentially illuminated with varying patterns, shown in Fig. 1(b). At each time the single-pixel detector records the total light intensity of the entire object scene, which is the inner product between the object image and the illumination pattern mathematically. After many illuminations with different patterns, the object image can be computationally reconstructed from the recorded single-pixel intensity sequence and the illumination patterns. The illumination light source in SPI is usually incoherent such as a light-emitting diode (LED). The principles, systems, algorithms and applications of SPI have been extensively investigated in many previous works [4-8].

In some previous works [9-13], a SPI system is utilized not only for capturing an object image but also for optical information processing such as optical image encryption, watermarking, and tracking. In this paper, a SPI system is designed for optical machine learning purpose without conventional image acquisition and reconstruction. It is assumed that the object image X contains totally N pixels, denoted by a vector of length N, X= $[x_1, x_2, ..., x_N]$ ($x_n$ denotes the intensity value of the $n_{th}$ pixel). The illumination pattern W= $[w_1, w_2, ..., w_N]$ has the same number of pixels. The recorded single-pixel intensity value y can be expressed as y=$w_1 x_1 + w_2 x_2 + \cdots + w_N x_N$. This SPI model concides with the single-layer perceptron (or linear classifier) model [14,15] in machine learning. In fact, the DNN system [1] is essentially a linear classifier as well. If the vector W is appropriately designed, the value y can indicate whether X belongs to one class or another. The optimal W can be determined from a large set of labeled training samples. For example, if the classification objective is to determine whether the image X belongs to a "dog" or a "cat". A set of different "dog" and "cat" images are prepared. The target output for each "dog" image is set to be T=1 and the target output for each "cat" image is set to be T=-1. All the images corresponding to different X and T values are employed to optimize W iteratively. Originally all the pixels in W are set to be random values. For a given X, the output y=WX, which is usually not equal to T (1 or -1). Then each element in W (or each pixel in the illumination pattern) is updated in the following way: $w_n' = w_n + r(T - y)x_n (1 <= n <= N)$, where r denotes the pre-defined learning rate. All the elements in W are iteratively updated in this way for each pair of X and T. Finally, the adaptively optimized W may yield outputs close to their correct target values (1 or -1) for most of the inputs X.

The MLSPI system can perform both binary classification and multiple-category classification. For example, if the objective is to recognize 10 different number digits (0, 1, 2, ..., 9), ten different illumination patterns are required and each illumination pattern is a binary classifier for each digit correspondingly. The first classifier is trained to output 1 when the input X is digit "0 " and output -1 when the input X is not "0" (any other number). The second classifier will output 1 when the input X is digit "1" and output -1 when the input X is not "1" and so on. Finally ten single-pixel intensity values are recorded and the one with the highest intensity indicates the input image most likely belongs to that corresponding digit number.

The MLSPI framework proposed in this paper and the DNN system proposed in the previous work [1] can both perform many linear machine learning tasks, such as digit classification. In a DNN system, the classification result is obtained by comparing the intensities in different sub-regions in the output imaging plane. In a MLSPI system, the classification result is obtained by comparing the intensities in a single-pixel value sequence, shown in Fig. 2.

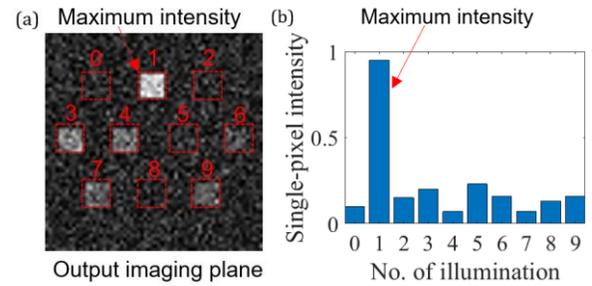

**Fig. 2. Different ways of displaying the classification result from the system output: (a) DNN; (b) MLSPI.**

**Table 1. Comparison between two kinds of optical machine learning systems**

|  | DNN | MLSPI |
|---|---|---|
| Illumination light | Coherent | Incoherent |
| Number of recordings | One | Multiple |
| Encoded parameters | Pixels in phase masks | Pixels in illumination patterns |
| Detectors | Pixelated sensor array | Single-pixel detector |
| Input format | Both intensity and phase | Only intensity |
| Display of classification and recognition results | Different sub-regions in the output imaging plane | A sequence of intensity values |
| Precision requirement in experiments | High | Relatively low |
| To be programmable | Relatively hard | Easy |

However, there are several differences between these two kinds of optical systems and each one has its own pros and cons, shown in Table 1. The proposed MLSPI system in this work has one major advantage that it can work under incoherent light illumination and has a relatively low alignment precision requirement in practical optical experiments. The DNN system can process complex-amplitude input signals containing both amplitude and phase such as orbital angular momentum modes [16,17] while the MLSPI system can only process real-intensity input signals. In addition, the MLSPI system requires multiple recordings, which will reduce the processing speed. But multiple illuminations and recordings can be implemented simultaneously if wavelength multiplexing and polarization multiplexing are used in the MLSPI system. The output result in a DNN system only needs to be captured once regardless of how many categories to be classified in the task. As a computing device, another key aspect is whether it is programmable. The illumination patterns projected in a MLSPI system can be freely and easily encoded for different machine learning tasks. However, the phase masks in the work [1] are fabricated with 3D printing, which are fixed and non-programmable once after fabrication.

In this work, a MLSPI system is implemented for the classification of handwritten digits (MINST dataset [18]), traffic sign images (BelgiumTS dataset [19]) and fashion images (Fashion MINST dataset [20]). Each dataset consists of training images and testing images, which are evenly divided into ten different categories. The ten illumination patterns are designed with the training algorithm described above. It shall be noted that the optimal weighting coefficients contain both positive and negative values. But the projection device can only project illumination patterns with positive pixel values in SPI. Consequently an off-set value needs to be added to the pixel values in the optimized illumination patterns to ensure that every pixel has a positive value. Then the pixel values in the patterns are normalize to [0 1]. The pixels values below certain lower bound and above certain higher bound maybe truncated to enhance the contrast. Some image examples and the corresponding designed illumination patterns for three datasets are shown in Fig. 3. It can be observed that the patterns contain extracted visual features for each category of images. In the design of illumination patterns, random variations (e.g. translational shift, rotation and scaling) can be added to the training images to enhance the system robustness in the classification.

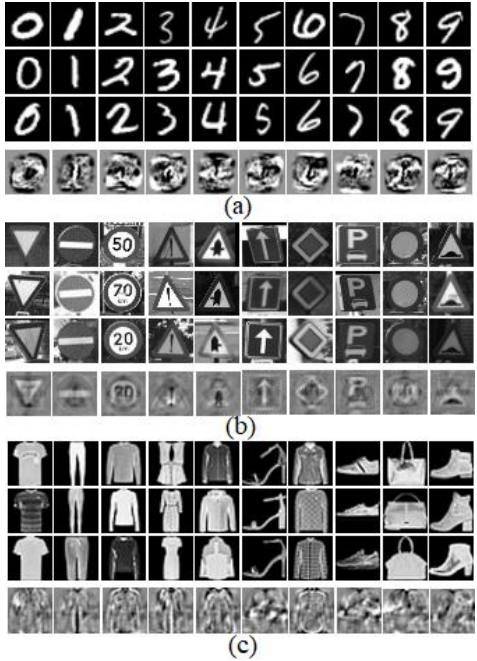

**Fig. 3. Image examples (first three rows) and designed illumination patterns in our scheme (last row) for each dataset: (a)MINST handwritten digits; (b)BelgiumTS traffic sign images; (c)Fashion-MINST images.**

In the simulation, the ten illumination patterns are projected onto randomly selected testing images and the single-pixel intensity values are recorded sequentially. The classification result for each object image is determined based on the maximum intensity value among the ten values in the recorded single-pixel intensity sequence. The simulated classification accuracies for each dataset are shown in Table 2 and most images can be correctly recognized in the simulation. For example, the classification accuracy for the MINST dataset (85.83%) approximately agrees with the accuracy of a linear classifier (around 88%) reported in the past literatures [18], where mean square error is the loss function in training. The classification accuracy in our work is not very high compared with some state-of-the-art works combining SPI with digital machine learning algorithms [21,22], where it may reach to nearly 100% accuracy [22]. However, the MLSPI system in this work is implemented in an all-optical manner instead of an optical-electronic hybrid system [21-24]. The output results of SPI are not further post-processed by any digital algorithm (e.g. a deep convolutional network) in this work. For such a full-optical system, the computing speed can be very fast and it remains unchanged as the object image size grows. The system may work under very weak light conditions by a single-photon detection, which requires very low power consumption.

**Table 2. Classification accuracy for three datasets with our proposed MLSPI system**

|  | Number digits | Traffic sign images | Fashion images |
|---|---|---|---|
| No. of training images | 20000 | 1000 | 20000 |
| No. of testing images | 120 | 100 | 120 |
| No. of categories | 10 | 10 | 10 |
| Classification accuracy (simulation) | 85.83% | 92% | 82.5% |
| Classification accuracy (experiment) | 84.17% | 90% | 75.83% |

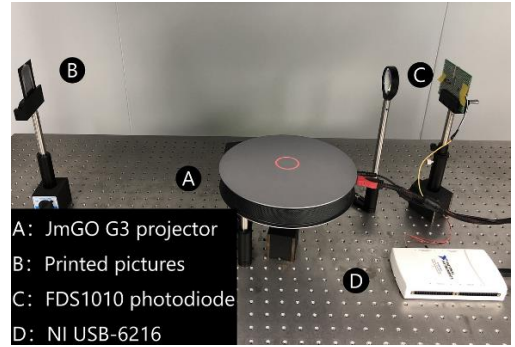

**Fig. 4. Optical setup for our proposed MLSPI system (A projects patterns to B. C will collect the total light intensity through a lens and the intensity data recorded by C is acquired by D).**

An optical experiment is carried out to further verify our proposed scheme. Each object image is printed on a paper card and sequentially illuminated by the ten patterns projected by a JmGO G3 projector. The size of each pixel in the object image or illumination pattern is approximately 0.2cm×0.2cm. The single-pixel intensity values are recorded by a Thorlabs FDS1010 photodiode detector and a NI USB-6216 data acquisition card. The optical setup is shown in Fig. 4. Due to noise and distortion, the classification accuracies in the experiment are slightly lower but close to the ones in the simulation, shown in Table 2 and Fig. 5. The confusion matrix tables in Fig. 5 show the distributions of predicted labels for each category. Most predicted labels are distributed along the diagonal line and match with the true labels. The results indicate that our proposed system can optically perform the classification task precisely for most testing objects.

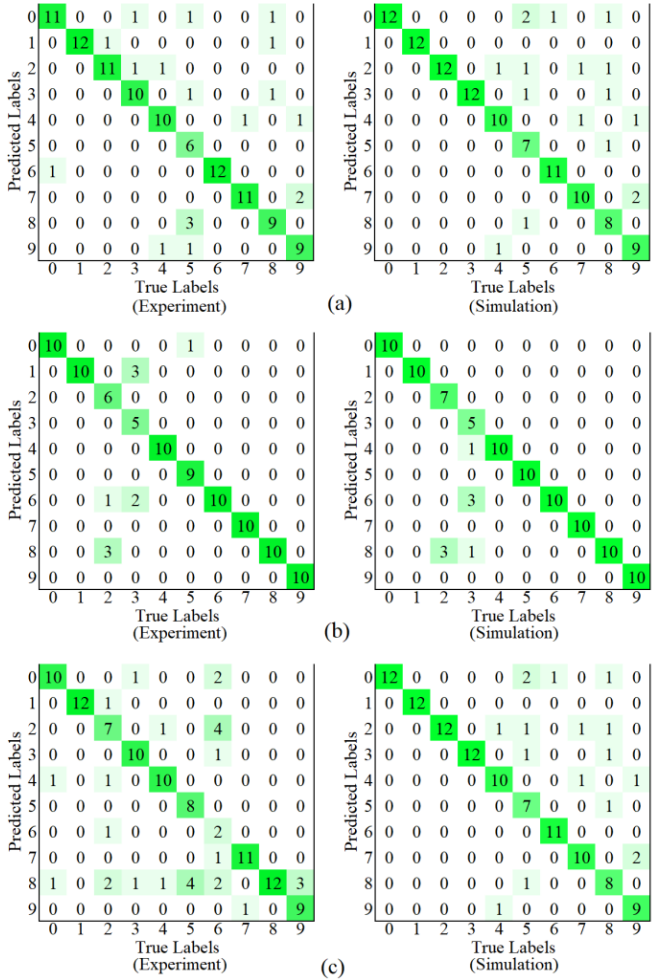

**Fig. 5.** Confusion matrix between the true labels and predicted labels for (a) MINST handwritten digits; (b) BelgiumTS traffic sign images; (c) Fashion-MINST images in the simulation (right) and experiment (left).

The classification accuracy of our proposed linear system for more complicated object images will be degraded. For example, it is 38.66% for the CIFAR-10 dataset [25] in the simulation. In future works, a hybrid optical-electronic design of MLSPI system with nonlinearities, like the previous work [26], will be investigated. A compact digital nonlinear processor can be employed to process the recorded data and optimize the illumination patterns in real time. Our proposed framework will be investigated in other systems such as Fourier ptychography [27] as well.

In summary, this paper proposes an optical machine learning framework based on single-pixel imaging (MLSPI). The MLSPI system can perform linear pattern recognition tasks such as number classification optically, verified by simulation and experimental results. Compared with the optical diffractive neural network (DNN) system proposed in the previous work [1], MLSPI has advantages of working under incoherent lighting, lower experimental complexity and being easily programmable.

**Funding.** National Natural Science Foundation of China (61805145, 11774240); Leading talents of Guangdong province program (00201505); Natural Science Foundation of Guangdong Province (2016A030312010).

**Acknowledgment**. We thank Prof. Zibang Zhang and Dr. Xiang Li in Jinan University (Guangzhou, China) for assistance.

**Disclosures**. The authors declare no conflicts of interest.